\documentclass[letterpaper]{article} 
\usepackage{aaai25}  
\usepackage{times}  
\usepackage{helvet}  
\usepackage{courier}  
\usepackage[hyphens]{url}  
\usepackage{graphicx} 
\usepackage{natbib}  
\usepackage{caption} 
\usepackage{algorithm}
\usepackage{color}
\usepackage{amsfonts} 
\usepackage{amsmath}
\usepackage{subcaption}
\usepackage[capitalise]{cleveref}
\usepackage{amsthm}
\DeclareMathAlphabet{\mymathbb}{U}{BOONDOX-ds}{m}{n}

\usepackage{algorithm}
\usepackage{algpseudocode}
\usepackage{booktabs} 
\theoremstyle{definition}

\usepackage{multirow}
\usepackage{makecell}
\usepackage{newfloat}
\usepackage{listings}
\DeclareCaptionStyle{ruled}{labelfont=normalfont,labelsep=colon,strut=off} 
\floatstyle{ruled}
\newfloat{listing}{tb}{lst}{}
\floatname{listing}{Listing}
\title{Navigating Towards Fairness with Data Selection}
\author{
    Yixuan Zhang\textsuperscript{\rm 1},
    Zhidong Li\textsuperscript{\rm 2},
    Yang Wang\textsuperscript{\rm 2},
    Fang Chen\textsuperscript{\rm 2},
    Xuhui Fan\textsuperscript{\rm 3},
    Feng Zhou\textsuperscript{\rm 4,5}\thanks{Corresponding author.}
}
\affiliations{
    \textsuperscript{\rm 1}School of Statistics and Data Science, Southeast University, China\\
    \textsuperscript{\rm 2}Data Science Institute, University of Technology Sydney, Australia\\
    \textsuperscript{\rm 3}School of Computing, Macquarie University, Australia\\
    \textsuperscript{\rm 4}Center for Applied Statistics and School of Statistics, Renmin University of China, China\\
    \textsuperscript{\rm 5}Beijing Advanced Innovation Center for Future Blockchain and Privacy Computing, China\\

    zh1xuan@hotmail.com, \{zhidong.li, yang.wang, fang.chen\}@uts.edu.au, xuhui.fan@mq.edu.au, feng.zhou@ruc.edu.cn
}

\usepackage{bibentry}

\begin{document}

\maketitle

\begin{abstract}
Machine learning algorithms often struggle to eliminate inherent data biases, particularly those arising from unreliable labels, which poses a significant challenge in ensuring fairness. Existing fairness techniques that address label bias typically involve modifying models and intervening in the training process, but these lack flexibility for large-scale datasets. To address this limitation, we introduce a data selection method designed to efficiently and flexibly mitigate label bias, tailored to more practical needs. Our approach utilizes a zero-shot predictor as a proxy model that simulates training on a clean holdout set. This strategy, supported by peer predictions, ensures the fairness of the proxy model and eliminates the need for an additional holdout set, which is a common requirement in previous methods. Without altering the classifier's architecture, our modality-agnostic method effectively selects appropriate training data and has proven efficient and effective in handling label bias and improving fairness across diverse datasets in experimental evaluations.
\end{abstract}

%

\section{Introduction}
\label{sec:intro}
Fairness is a critical and essential problem in real-world applications. In recent years, it has attracted great attention, especially in high-stake domains such as finance~\citep{credit_example,risk_prediction,loan_approval}, law~\citep{compas,human_bias_compas}, recruiting~\citep{recruiting,employer}, school admissions~\citep{school_admission} and medicine~\citep{medical}. Although many fairness-aware learning methods have been proposed recently, they often assume that the data collected for training is representative of the true data distribution. However, these methods are still validated on so-called ``clean data," which neglects the impact of \emph{label bias}.

Recent studies~\citep{gpl,Dai2020LabelBL,pmlr-v206-zhang23g,JMLR:fair_corrupted} have increasingly focused on the adverse impacts of label bias and have proposed methodologies to improve fairness in this setting. These efforts involve adjusting models and intervening in the training process by considering the amount of bias present in labels during fair learning. The objective of these methods is to develop a fair labeling function that reflects the true underlying distribution, thereby improving the robustness of the fairness techniques. However, when handling large-scale datasets, these methods often lack flexibility due to the necessity of training with the complete dataset, which can slow down model convergence and substantially increase training costs.

To meet the needs of complex machine learning systems in practical applications, employing data selection methods to filter out a subset of useful data for training proves to be an effective solution. Apart from efficiency improvement, data selection techniques have been very effective in mitigating the impact of noisy data. These methods typically prioritize training with either difficult~\citep{DBLP:journals/corr/LoshchilovH15,pmlr-v80-katharopoulos18a,DBLP:journals/corr/jiang} or easy samples~\citep{Curriculum/10.1145/1553374.1553380} based on the training loss. Nevertheless, such a singular filtering strategy limits the ability to handle the diversity of real-world situations since the difficulty of samples often arises from incorrect annotations, inherent ambiguity, or atypical patterns~\citep{pmlr-v162-mindermann22a,deng2023towards}. To overcome this limitation, \citet{pmlr-v162-mindermann22a} introduces a new data selection criterion, the reducible holdout loss selection (RHO-LOSS), based on the impact on generalization loss, which further prevents the selection of redundant or noisy samples.

Inspired by this new selection criterion, this paper aims to extend this method to the field of fairness. We revisit the derivation of RHO-LOSS, aligning the predicted posterior distribution of selected samples with the fair data distribution. Instead of relying on a holdout set and training an auxiliary validation model, we establish a more accurate approximation by incorporating zero-shot predictors from pre-trained models. This approach eliminates the need for an additional holdout set. To further prevent discrimination leakage in pre-trained models, we implement the peer prediction mechanism during training. This ensures fairness when using the zero-shot predictor to evaluate generalization loss.

We conduct comprehensive empirical evaluations on several benchmark datasets. The experiments on the image classification tasks demonstrate the effectiveness of our proposed data selection principle, which can adaptively select fair instances that are less impacted by label bias. Although not explicitly emphasized in this paper, we also offer a solution to reduce the impact of selection bias by resampling the selected data. It is worth noting that the proposed method is \emph{modality-agnostic}, as it only provides a principle of data selection and can be compatible with any log-likelihood or cross-entropy-based classifiers. Meanwhile, it achieves faster convergence as we only select ``good" data points for model training. Our method is robust and achieves superior performance with respect to accuracy and fairness metrics under different bias settings. Our contributions are three-fold:
\begin{itemize}
    \item We propose a data selection principle that enables the learning of a fair labeling function by selecting fair and balanced instances in the training set to fix both label and selection bias. Notably, unlike most noisy label learning methods, our method does not require noise rate estimation.
    \item Our method is general, modality-agnostic, and compatible with any classifier or neural network based on log-likelihood or cross-entropy-based.
    \item Our method converges faster and achieves better test accuracy than alternative baselines.
\end{itemize}

\section{Preliminaries}
In this section, we briefly introduce the related background knowledge of online batch selection and present the concept of the data selection principle based on the data impact on the generalization loss, as proposed in \citet{pmlr-v162-mindermann22a}. 

Consider a dataset $\mathcal{D} = \{(x_i,y_i,s_i)\}^n_{i=1}$, where $x$ represents the non-sensitive features, $y\in\{0,1\}$ is the binary label, and $s\in \{0,1\}$ denotes the sensitive variables. Given a model parameterized by $\theta$, $f_\theta: \mathcal{X} \rightarrow \mathbb{R}^K$ ($K$ is the number of class), in online batch learning, a batch $B_t$ of size $N_B$ is drawn from the training dataset $\mathcal{D}$ at each training step $t$. The objective of online batch selection is to pick samples from $B_t$ based on specific ranking criteria and use them to construct a smaller batch $b_t$ with size $N_b$ for model updates. The ranking function represents the criteria of which data should be selected, and previous methods usually pick the ``hard'' data points based on the ranking of training loss from highest to lowest. Then, the common gradient descent is performed to minimize the loss using $b_t$, which is denoted as $\sum^{N_b}_{i=1}L(y_i,f(x_i))$. By iteratively doing this, we can select data points that minimize the loss of training set. However, this kind of selection method will tend to pick redundant data points or outliers since it focuses on the highest training loss and lack the flexibility due to the simplistic selection criteria.

In contrast, \citet{pmlr-v162-mindermann22a} established selection criteria based on the data impact on the model's generalization loss, thus effectively addressing the limitations of previous methods. 
To revisit this approach, we follow the framework of online batch selection. Let $\mathcal{D}_t$ represent the observed data prior to training step $t+1$. Given a sample $(x,y)$ drawn from batch $B_{t+1}$, we assume, for simplicity, that only one data point is selected at a time. If this data point is chosen, the updated predictive distribution in a Bayesian view will be $p(y'\mid x',\mathcal{D}_{t} \cup (x,y))$. This distribution should ideally align with the true data-generating distribution $\dot{p}(x',y')$, and to achieve this goal, one nature way is to minimize the KL divergence between the predictive distribution and the data-generating distribution, $\mathbb{E}_{\dot{p}(x')}KL[\dot{p}(y'\mid x') || p(y' \mid x', \mathcal{D}_{t}\cup (x,y)]$, which can be equivalently expressed as: 
\begin{equation}
    - \mathbb{E}_{\dot{p}(x',y')}[\log p(y'\mid x', \mathcal{D}_t \cup (x,y)]+\text{const.},
\end{equation}
where the $\text{const.}$ denotes the negative entropy of the data generating distribution, and is agnostic to the optimization. In order to evaluate the impact of the selected data point on the generalization loss, we utilize the holdout samples $\mathcal{D}^* = \{(x_i^*,y_i^*)\}_{i=1}^m$ from the data-generating distribution $\dot{p}(x',y')$. By leveraging the Monte Carlo to approximate expectations with empirical averages, the aim now becomes:
\begin{equation}
\label{eq: log_likelihood_of_holdout}
    \underset{(x,y)\in B_{t+1}}{\text{arg max}}\frac{1}{m}\sum^m_{i=1}[\log p(y_i^*\mid x_i^*,\mathcal{D}_t \cup (x,y))],
\end{equation}
and this optimization is equivalent to select a data point $(x,y) \in B_{t+1}$ that mostly maximizes $\log p(\mathcal{D}^* \mid \mathcal{D}_t \cup (x,y))$, which corresponds to the generalization loss. By applying Bayes' rule, we obtain:
\begin{equation}
\begin{aligned}
    p(\mathcal{D}^*\mid \mathcal{D}_t \cup (x,y)) = \frac{p(y\mid x, \mathcal{D}^*, \mathcal{D}_t)}{p(y\mid x,\mathcal{D}_{t})}p(\mathcal{D}^*\mid \mathcal{D}_t).
\end{aligned}
\end{equation}
The term $p(\mathcal{D}^*\mid \mathcal{D}_t,x)$ becomes $p(\mathcal{D}^*\mid \mathcal{D}_t)$ is due to a single data point $x$ does not influence the model's update performance. Approximating $p(y\mid x, \mathcal{D}^*, \mathcal{D}_t)$ with $p(y\mid x,\mathcal{D}^*)$ and dropping the term $p(\mathcal{D}^*\mid \mathcal{D}_t)$ (independent of $(x,y)$), the final approximated tractable selection function of RHO-LOSS is given by:
\begin{equation}
\begin{aligned}
\label{eq:apprx_rho_loss}
&\underset{(x,y)\in B_{t+1}}{\text{arg max }}\log p(y\mid x,\mathcal{D}^*)- \log p(y\mid x, \mathcal{D}_t)\\
=&\underset{(x,y)\in B_{t+1}}{\text{arg max }} L[y\mid x,\mathcal{D}_t] - L[y\mid x,\mathcal{D}^*]. 
\end{aligned}
\end{equation}
The first term, $L[y\mid x,\mathcal{D}_t]$, represents the training loss using model trained on the training set $\mathcal{D}_t$, while $L[y \mid x,\mathcal{D}^*]$ is the irreducible holdout loss using model trained on the holdout set $\mathcal{D}^*$. Thus, the aim of selecting a data point $(x,y)\in B_{t+1}$ that mostly maximize the log-likelihood on the holdout set (\cref{eq: log_likelihood_of_holdout}) can be approximated by \cref{eq:apprx_rho_loss}.


\section{Method}

Building on the concept of RHO-LOSS, we now turn our attention to how it connects with fairness and how label bias impacts data selection.  

\subsection{Refining Selection Principle for Fairness}
We begin by revisiting the derivation of RHO-LOSS and incorporating fairness principles (demographic parity\footnote{We use demographic parity as an illustration here, other fairness metrics are introduced in the experiment.}) to refine its formulation. Consider a holdout dataset $\mathcal{D}^* = \{x_i^*,y_i^*,s_i^*\}^m_{i=1}$, where all samples are generated from a fair distribution $p(x,y,s) = p(y\mid x)p(x)p(s)$, meaning labels are not influenced by sensitive attributes. While fairness has various interpretations~\citep{Barocas2018FairnessAM}, we follow a common assumption in fairness literature and it is worth noting that, such a fair distribution is an idealized condition and is not observed in practice. Our objective now is to select a sample $(x,y)$ from batch $B_{t+1}$, belonging to the demographic group defined by $s$, ensuring that the updated predictive distribution closely aligns with the fair data distribution. Similar to the derivation in RHO-LOSS, by expanding the KL expression and applying Monte Carlo estimation using additional holdout samples, the optimization problem becomes: 

\begin{equation}
\label{eq: optimzation_goal}
    \underset{(x,y)\in B_{t+1}}{\text{arg max}}\sum_s p(s)\left[\log p(\mathcal{D}^* \mid \mathcal{D}_t\cup (x,y),s)\right]. 
\end{equation}
We estimate $p(s)$ using $\frac{C_s}{m}$, where $C_s$ represents the number of samples belonging to demographic group $s$ under fair distribution. For demographic group $s$, using Bayes' rule, the updated predictive distribution is as follows:
\begin{equation}
\label{eq: bayes_updated}
    \begin{aligned}
        p(\mathcal{D}^*\mid \mathcal{D}_t \cup (x,y),s) = \frac{p(y\mid x,\mathcal{D}^*, \mathcal{D}_{t},s)}{p(y\mid x, \mathcal{D}_{t},s)} p(\mathcal{D}^* \mid \mathcal{D}_{t}).
    \end{aligned}
\end{equation}
By plugging \cref{eq: bayes_updated} into \cref{eq: optimzation_goal} and omitting the term that is irrelevant to the selected sample, we reformulate the objective as:
\begin{equation}
\label{eq: optimzation_new}
\begin{aligned}
    \underset{(x,y)\in B_{t+1}}{\text{arg max}}\sum_s \frac{C_s}{m}&[\log p(y\mid x,\mathcal{D}^*,\mathcal{D}_{t},s)\\ &-\log p(y\mid x, \mathcal{D}_{t},s)].
\end{aligned}
\end{equation}
The second term in \cref{eq: optimzation_new} is straightforward to compute, but the first term is challenging to estimate because it involves both the training and holdout data. To address this, RHO-LOSS approximates the term with $\log p(y\mid x,\mathcal{D}^*)$, while \citet{deng2023towards} approximates it by finding a lower bound. We adopt the lower bound approximation, which is expressed as follows: 
\begin{equation}
\label{eq:lowe_bound_term1}
\log p(y\mid x,\mathcal{D}^*,\mathcal{D}_{t},s) \ge \mathbb{E}_{p(\theta \mid \mathcal{D}^*)}\log p(y\mid x,\theta,s), 
\end{equation}
where $\theta$ is the model parameter. 
For completeness, we include the full derivation in the Appendix. 
Substituting \cref{eq:lowe_bound_term1} into \cref{eq: optimzation_new}, the resulting objective is as follows: 
\begin{equation}
\label{eq: optimzation_final}
\begin{aligned}
    \underset{(x,y)\in B_{t+1}}{\text{arg max}} \sum_s \frac{C_s}{m} &\left[\alpha \mathbb{E}_{p(\theta \mid \mathcal{D}^*)} \log p(y \mid x, \theta, s) \right. \\
    & \left. - \log p(y \mid x, \mathcal{D}_t, s)\right],
\end{aligned}
\end{equation}
where $\alpha\in [0,1]$ is a scaling factor that determines the trade-off between the original inequality and the lower bound. 

Although we obtain this lower bound to approximate the first term in~\cref{eq: optimzation_new}, it cannot be directly calculated. To further make it tractable, we build on the framework outlined by~\citet{deng2023towards} to approximate this lower bound using zero-shot predictors as a proxy for the validation model in \citet{pmlr-v162-mindermann22a}, thereby eliminating the need to gather additional holdout data. Zero-shot predictors have demonstrated promising transfer performance across a wide range of downstream tasks due to training on extensive datasets. Consequently, we use the following approximation: 
\begin{equation}
\label{eq:zero_shot_approximation}
    \mathbb{E}_{p(\theta \mid \mathcal{D}^*)}\log p(y\mid x,\theta,s) \approx \log p(y\mid \Tilde{f}(x),s),
\end{equation}
where $\Tilde{f}$ represents a zero-shot predictor derived from a pre-trained model used as a validation model. The approximation is considered reasonable because the expectation is taken over the posterior of $\theta$, and it is assumed that the dataset used to train the zero-shot predictor is sufficiently large, resulting in a very narrow posterior distribution. Consequently, we can directly extract the posterior mean, represented by $\Tilde{f}$, and substitute it into $\log p(y\mid x,\theta;s)$ as an effective approximation of $\mathbb{E}_{p(\theta \mid \mathcal{D}^*)}\log p(y\mid x,\theta;s)$. Then, substitute \cref{eq:zero_shot_approximation} into \cref{eq: optimzation_final}, the selection function with fairness considerations can be reformulated as:
\begin{equation}
\label{eq:objective}
    \underset{(x,y)\in B_{t+1}}{\text{arg max}}\sum_{s} \frac{C_s}{m} \left [ L[y\mid x, \mathcal{D}_t,s] - \alpha L[y,\Tilde{f}(x),s] \right ].
\end{equation}
The final objective is structured similarly to RHO-LOSS: the first term denotes the training loss, while the second term approximates the holdout loss via a zero-shot predictor.

\subsection{Fair Data Selection with Peer Prediction Mechanism}

In this section, we present our revised selection approach to enhance fairness. Previous methods rely on a clean holdout set for training validation models, which is not always feasible in practice. Instead, we use zero-shot predictors as a proxy model, eliminating the need for a holdout set (see \cref{eq:objective}). Despite the pre-trained model's robust ability to effectively extract and utilize fundamental patterns from large datasets, it may still inherit and propagate label bias, failing to adequately reflect the data-generating distribution of the current task, especially in terms of fairness. This misalignment can lead to persistent discrimination in the proxy model's approximations, mainly because most pre-trained models neglect to emphasize fairness during training, resulting in inherent biases. Even models developed with fairness considerations can introduce biases in downstream tasks due to distribution shifts~\citep{jiang2023chasing,schrouff2022diagnosing,fair_incremental_learning}. To address these issues and tackle inherent label bias, we incorporate a peer prediction mechanism into our zero-shot predictor.

\paragraph{Peer Prediction Mechanism} The correlated agreement type~\citep{dasgupta_peer_prediction,shnayder_peer_prediction} of peer prediction mechanism involves two agents: one that provides noisy labels and another that mimics the Bayes optimal classifier. \citet{liu2020peer} crafted a scoring function that encourages truthful reporting by ensuring that the optimal classifier maximizes its score with accurate predictions. By minimizing the negative scoring function as a loss (referred to as \textit{peer loss}), the resulting classifier closely approximates the Bayes optimal classifier, effectively addressing challenges posed by noisy labels. The loss function is formulated as: $L[y_i \mid x_i] - \gamma L[y_{i_2}\mid x_{i_1}]$, where $x_{i_1}$ and $y_{i_2}$ are independently sampled from the training set, excluding the $(x_i,y_i)$ pair, and $\gamma \in [0,1]$ is a parameter that makes the loss robust to imbalanced labels. 

\paragraph{Final Fair Data Selection Objective} Drawing on the principles of the peer prediction mechanism, we implement a similar strategy to eliminate label bias and hence ensure fairness. We define $\mathcal{D}_{s}$ and $\mathcal{D}_{s'}$ as the subsets containing sensitive information for $S=s$ and $S=s'$, respectively. In a fair setting, predictions should be independent of $s$. To achieve this, we modify the original peer loss by sampling $x_{i_1}$ from $\mathcal{D}_{s}$ and $y_{i_2}$ from $\mathcal{D}_{s'}$, and the corresponding random variables are the pair of $(X_{i_1}, Y_{i_2})$. The intuition behind this design is that we assume the bias exists, and we construct such cross-group pairings to create a ``biased" version. This setup allows us to estimate the bias rate conditioned on different demographic groups using the peer loss function. Since this loss function involves pairing two randomly selected instances, it inherently contains some randomness. To stabilize the results, we calculate the expectation across these randomly selected instances, and the equivalent expectation version is (details in the Appendix): 
\begin{equation}
\begin{aligned}
    \label{eq:expectation_peer}
    L[y_i,\Tilde{f}(x_i),s] - \gamma \mathbb{E}_{Y\mid \mathcal{D}_{s'}}[L[Y,\Tilde{f}(x_i),s]]. 
\end{aligned}
\end{equation}
Combining \cref{eq:expectation_peer} into \cref{eq:objective} and simplifying \cref{eq:objective}, with the weights $\frac{C_s}{m}$ summing to 1, we obtain the final selection function: 
\begin{equation}
\begin{aligned}
\label{eq:final_objective}
    \underset{(x,y)\in B_{t+1}}{\text{arg max}}   &L[y\mid x, \mathcal{D}_t,s] + (1-\alpha) L[y,\Tilde{f}(x),s] \\ 
    &- \gamma \mathbb{E}_{Y\mid \mathcal{D}_{s'}}[L[Y, \Tilde{f}(x),s]].
\end{aligned}
\end{equation}

\paragraph{Resampling to Deal with Selection Bias} 
In addition to addressing label bias, our implementation also tackles selection bias, which contributes to label imbalance. Assuming selection bias impacts the statistical independence between demographic groups $S$ and target labels $Z$~\citep{kamiran_calders_2012}, we resample based on the discrepancies between the actual counts of individuals in demographic group $s=i$ with clean label $z=j$ ($C_{s,z}$) and their expected counts ($\mathbb{E}[C_{s,z}]$). We estimate $\mathbb{E}[C_{s,z}]$ using $p(s)p(z)N$, where $p(s)$ and $p(z)$ are the empirically measured probabilities. To correct imbalances, we upsample subgroups where $C_{s,z} < \mathbb{E}[C_{s,z}]$ and downsample those where $C_{s,z} > \mathbb{E}[C_{s,z}]$. Although the clean label $Z$ is unobserved, we assume that the labels of selected instances are fair and treat them accordingly. This resampling strategy is applied to each selected batch to ensure data balance.

\paragraph{Implementation} The training process is outlined in \cref{alg:fairfilter}. We employ a zero-shot predictor with a derived expectation version of the adapted peer loss as a surrogate model to simulate the estimated loss on the holdout set. For each instance in batch $B_{t+1}$, we evaluate \cref{eq:final_objective} and select the top-$N_b$ instances to form a smaller batch $b_{t+1}$. To address selection bias, we further pick samples based on the severity of selection bias to ensure balance among different demographic groups. We then update the parameters $\theta$ of the target model using this selected data.

\begin{algorithm}
  \caption{Fair data selection to address label bias issue}\label{alg:fairfilter}
  \begin{algorithmic}[1]
   \State \textbf{Input:} training set $\mathcal{D}$, $N_b$, $N_B$, $T$, $\alpha$, $\gamma$, zero-shot predictor $\Tilde{f}$ and a target model $f_\theta$.
      \State Initialize $\theta_0$.
    \For{\texttt{$t$ in $0,\cdots,T$}}
    \State Randomly select $N_B$ instances to construct $B_{t+1}$;
    \State For each sample $(x_i,y_i,s_i)$ in $B_{t+1}$, estimate and compute the objective in \cref{eq:final_objective};
    \State Select top-$N_b$ samples to construct $b_{t+1}$;
    \State Drop instances from $b_{t+1}$ if $C_{s,z} > \mathbb{E}[C_{s,z}]$, otherwise, bootstrap; 
    \State Perform gradient descent and update $\theta$ with re-sampled data. 
    \EndFor
  \end{algorithmic}
\end{algorithm}

\paragraph{Why the fair selection principle pick instances less influenced by label bias?} Our data selection method integrates a zero-shot predictor with a peer prediction mechanism to address label bias, using the derived loss as a validation measure against a fair distribution. By evaluating each data point’s impact on this validation loss, we avoid instances likely discriminated by sensitive information and exclude irrelevant outliers. This approach also conceptually relates to the Query By Committee (QBC) algorithm~\citep{qbc_1992,QBC_1997}, which selects examples based on classifier disagreement (here, between the current model and $\Tilde{f}$). Following \citet{cheng2020learning,pmlr-v157-zhang21d} (detailed in the Appendix), we demonstrate that our method effectively promotes fairness and mitigates label bias through a breakdown of \cref{eq:expectation_peer}:

\begin{equation}
\begin{aligned}
\label{eq: analysis_decomposition}
  &\mathbb{E}_{\Tilde{D}}\left [L[Y,\Tilde{f}(X),S]- \gamma \mathbb{E}_{Y\mid \mathcal{D}_{s'}}L[Y, \Tilde{f}(x),s]\right ]\\
   &= \underbrace{\mathbb{E}_{D}\Delta_s L[Y,\Tilde{f}(X),S]}_\text{fair model}\\
   &+ \underbrace{\sum_{j\in [K]} \sum_{i \in [K]} \sum_{s\in [S]} P(S=s)P(Z=i)[\Delta_{D_u} L[j,\Tilde{f}(X),s]]}_\text{Penalty on noisy loss}\\
   &\underbrace{-\gamma \sum_{j\in [K]}  p_j^s \left [\mathbb{E}_{D_{X\mid S=1}}L[j,\Tilde{f}(X)]-\mathbb{E}_{D_{X\mid S=0}}L[j,\Tilde{f}(X)]\right ],}_\text{Penalty for the disagreement between demographic groups}
    \end{aligned}
\end{equation}
where $\Tilde{D}$ denotes the observed distribution, which contains label bias, while $D$ represents the underlying clean fair distribution. $\Delta_s = 1-P(Y=1\mid Z=0,S=s ) - P(Y=0\mid Z=1,S=s)$, $p_j^s = P(Y=j, S=0) - P(Y=j, S=1)$, $T_{ij}^s = P(Y=j\mid Z=i,S=s)$, $\Delta_{D_u} = \mathbb{E}_{D\mid Z=i, S=s}(U_{ij}^s - \gamma P(Y=j\mid S=s) )$ and $U_{ij}^s = T_{ij}^s$ if $i \neq j$, and $U_{ij}^s =  T_{jj}^s - \mathbb{E}_{D\mid Z=j,S=s}T_{jj}^s$ if $i=j$. The equation's first term captures the model's clean loss on the fair distribution. The second term adds penalties for noisy losses, adjusting labels to account for observed demographic disparities. The third term imposes penalties for demographic discrepancies, ensuring fair performance across all groups.

\begin{table*}[!t]
    \centering
    \resizebox{0.9\linewidth}{!}{
    \begin{tabular}{c c c c c c c c c}
    \toprule
     Method/Dataset & \multicolumn{2}{c}{LFW+a(0.2)}&\multicolumn{2}{c}{LFW+a(0.4)} & \multicolumn{2}{c}{CelebA(0.2)} & \multicolumn{2}{c}{CelebA(0.4)} 
     \\
     \midrule
     & ACC& $\Delta_{\text{DP}}$ & ACC& $\Delta_{\text{DP}}$&ACC& $\Delta_{\text{DP}}$&ACC& $\Delta_{\text{DP}}$ \\
     \midrule
     CLIP  & 90.8 & 0.11 & 83.2 & 0.21 & 62.2 & 0.60 & 61.6 & 0.61 \\
     \midrule
     Grad Norm & 77.5$_{\pm0.7}$ & 0.02$_{\pm 0.01}$ & 75.6$_{\pm 0.8}$ & 0.07$_{\pm 0.01}$ & 76.4$_{\pm 2.8}$ & 0.28$_{\pm 0.07}$ & 74.0$_{\pm 1.9}$ & 0.26$_{\pm 0.02}$\\
     Grad Norm IS & 82.6$_{\pm 1.4}$ & 0.43$_{\pm 0.13}$ & 75.4$_{\pm 0.2}$ & 0.78$_{\pm 0.01}$ & 82.6$_{\pm 2.1}$ & 0.40$_{\pm 0.14}$ & 75.5$_{\pm 3.4}$ & 0.42$_{\pm 0.07}$\\
     Uniform & 89.0$_{\pm 0.9}$ & 0.03$_{\pm 0.01}$ & 79.8$_{\pm 5.1}$ & 0.07$_{\pm 0.04}$ & 82.0$_{\pm 1.3}$ & 0.29$_{\pm 0.08}$ & 81.0$_{\pm 3.1}$ & 0.31$_{\pm 0.09}$ \\
     RHO-LOSS & 89.5$_{\pm 0.5}$ & 0.06$_{\pm 0.02}$ & 83.9$_{\pm 1.8}$ & 0.08$_{\pm 0.20}$ & 80.5$_{\pm 0.9}$ & 0.22$_{\pm 0.07}$ & 79.7$_{\pm 1.7}$ & 0.28$_{\pm 0.04}$\\
     \midrule
     Ours-s & 90.0$_{\pm 1.3}$ & 0.02$_{\pm 0.01}$ & 86.2$_{\pm 0.7}$ & 0.08$_{\pm 0.05}$ &  85.5$_{\pm 0.6}$ & \textbf{0.21$_{\pm 0.02}$} & 84.5$_{\pm 0.6}$ & 0.21$_{\pm 0.01}$\\
     Ours & \textbf{90.9$_{\pm 0.6}$} & \textbf{0.01$_{\pm 0.00}$} & \textbf{88.7$_{\pm 0.7}$} & \textbf{0.04$_{\pm 0.01}$} & \textbf{86.5$_{\pm 0.6}$} & \textbf{0.21$_{\pm 0.02}$} & \textbf{85.2$_{\pm 1.7}$} & \textbf{0.20$_{\pm 0.01}$} \\
     \bottomrule
    \end{tabular}}
    \caption{Test accuracy (\%) and fairness violation ($\Delta_{\text{DP}}$) on CelebA and LFW+a with various symmetrical label bias amount of 20\% and 40\%. We report results in the format of mean $\pm$ standard deviation.}
    \label{tab:acc_compare}
\end{table*}

\begin{table*}[ht]
    \centering
    \resizebox{0.9\linewidth}{!}{
    \begin{tabular}{c c c c c c c c c}
    \toprule
     Method/Dataset & \multicolumn{2}{c}{LFW+a(0.2)}&\multicolumn{2}{c}{LFW+a(0.4)} & \multicolumn{2}{c}{CelebA(0.2)} & \multicolumn{2}{c}{CelebA(0.4)} 
     \\
     \midrule
     & p\% ratio & $\Delta_{\text{DEO}}$ & p\% ratio& $\Delta_{\text{DEO}}$&p\% ratio& $\Delta_{\text{DEO}}$&p\% ratio& $\Delta_{\text{DEO}}$ \\
     \midrule
     CLIP  & 88.9 & 0.03 & 77.4 & 0.07 & 38.3 & 0.32 & 37.3 & 0.33\\
     \midrule
     Grad Norm & 92.7$_{\pm 1.5}$ & 0.03$_{\pm 0.01}$ & 91.1$_{\pm 0.6}$ & - & 77.6$_{\pm 7.3}$ & 0.39$_{\pm 0.02}$ & 76.0$_{\pm 8.9}$ &  0.55$_{\pm 0.23}$\\
     Grad Norm IS & 90.4$_{\pm 2.6}$ & \textbf{0.01$_{\pm 0.01}$} & 90.2$_{\pm 0.5}$ & - &31.9$_{\pm 0.0}$ & - & 32.5$_{\pm 0.0}$ & -\\
     Uniform & 97.9$_{\pm 1.9}$ & \textbf{0.01$_{\pm 0.01}$} & 90.1$_{\pm 5.2}$ & \textbf{0.14$_{\pm 0.09}$} & 76.3$_{\pm 3.9}$ & 0.52$_{\pm 0.33}$ & 68.6$_{\pm 5.9}$ & 0.77$_{\pm 0.10}$  \\
     RHO-LOSS & 92.3$_{\pm 2.8}$ & 0.03$_{\pm 0.01}$ & 90.4$_{\pm 2.4}$ & - & 80.8$_{\pm 5.9}$ & 0.44$_{\pm 0.14}$ & \textbf{81.0$_{\pm 0.1}$} & 0.42$_{\pm 0.01}$\\
     \midrule
     Ours & \textbf{98.3$_{\pm 0.4}$} & \textbf{0.01$_{\pm 0.00}$}& \textbf{94.6$_{\pm 3.7}$} & \textbf{0.14$_{\pm 0.08}$} & \textbf{84.1$_{\pm 5.3}$} & \textbf{0.38$_{\pm 0.15}$} & 77.2$_{\pm 1.1}$ & \textbf{0.35$_{\pm 0.02}$}\\
     \bottomrule
    \end{tabular}}
    \caption{Other fairness measure with p\%-rule and $\Delta_{\text{DEO}}$ on CelebA and LFW+a with various symmetrical label bias amount of 20\% and 40\%. 
    `-' denotes the invalid measure of DEO due to low accuracy.}
\label{tab:other_fair_measure}
\end{table*}

\section{Experiment}
\label{sec:exp}
 In the subsequent sections, we first describe our experimental setup, covering datasets, baselines, and evaluation metrics. Next, we compare our methods against existing state-of-the-art data selection techniques across various image classification tasks (CelebFaces Attributes (\textbf{CelebA})~\citep{celeba} and modified Labeled Faces in the Wild Home (\textbf{LFW+a})~\citep{lfwa_usage}), considering different amounts of label bias. We examine our selection criteria through detailed ablation studies.

\textbf{Benchmark Datasets.} We evaluate the performance of our proposed method using two image datasets: CelebA and LFW+a. The CelebA dataset is utilized to discern the label HeavyMakeup, considering gender (``Female'') as the sensitive variable where biases have been noted towards female. In the LFW+a dataset, we augment each image with additional attributes like gender and race (same in CelebA), aiming to classify the identity's gender. The sensitive variable here is ``WavyHair'', where literature has shown a strong correlation regarding males. Each dataset is divided into training, validation, and test sets.

\textbf{Baselines.} To evaluate our method’s effectiveness and robustness, we compare our method with several selection methods on the image tasks. These include uniform sampling (\textbf{Uniform}), gradient norm selection (\textbf{Grad Norm}, which selects data points with high gradient norms) \citep{pmlr-v80-katharopoulos18a}, and gradient norm with importance sampling (\textbf{Grad Norm IS}) \citep{pmlr-v80-katharopoulos18a}. Additionally, we compare it with \textbf{RHO-LOSS} \citep{pmlr-v162-mindermann22a}. We implement two variants of our method: one includes resampling of $b_{t+1}$ (\textbf{Ours}), and the other does not (\textbf{Ours-s}).

\textbf{Evaluation Metrics.} We use accuracy to evaluate prediction performance and measure fairness violation with $\Delta_{\text{DP}}=\vert\mathbb{E}(\hat{Y}=1\mid S=1) - \mathbb{E}(\hat{Y}=1\mid S=0)\vert$. A lower $\Delta_{\text{DP}}$ indicates less fairness violation. We also conduct experiments on the difference of equal opportunity (DEO)~\citep{equal_opportunity}, which is defined as $\Delta_{\text{DEO}} = \vert\mathbb{E}(\hat{Y}=1\mid Y=1, S=1)-\mathbb{E}(\hat{Y}=1\mid Y=1, S=0)\vert$, and the p\%-rule, $\text{p\%} = \text{min}(\frac{P(\hat{Y}=1\mid S=0)}{P(\hat{Y}=1\mid S=1)}, \frac{P(\hat{Y}=1\mid S=1)}{P(\hat{Y}=1\mid S=0)})$. A lower $\Delta_{\text{DEO}}$ suggests less fairness violation, while a lower p\%-rule indicates higher fairness violation.

\textbf{Setup.} In our experiments addressing label bias, we introduce symmetrical label biases of 20\% and 40\%. We use the AdamW optimizer (learning rate 0.001, weight decay 0.01), we set a batch size of $N_b=32$ and a batch ratio $\frac{N_b}{N_B}=0.1$, consistent with the RHO-LOSS setup. For the LFW+a dataset, we employ ResNet-18 \citep{DBLP:journals/corr/HeZRS15}, and for CelebA, we use ResNet-50 across all methods, along with a zero-shot predictor based on CLIP-RN50. We vary $\alpha$ and $\gamma$ within the set $\{0.1, 0.3, 0.5, 0.7, 0.9\}$. Results are averaged over three random trials. All experiments are performed with GPUs (NVIDIA GeForce RTX 3090 with 86GB memory).

\subsection{Comparison Results}
Results are displayed in \cref{tab:acc_compare} for 20\% and 40\% bias amount. In the meantime, we report the fairness measure using p\%-rule and $\Delta_{\text{DEO}}$ in \cref{tab:other_fair_measure}. Our proposed method consistently demonstrates the highest accuracy and minimal fairness violation as bias increases. The outcomes illustrate the importance of addressing label bias to prevent heightened bias in the output. While other baselines work well with low bias, they are not robust when the bias amount increases. Interestingly, we find that gradient norm selection has the worst performance, even worse than uniform sampling, especially for large label bias amount. This phenomenon shows that selecting data by high variance will tend to pick ``dirty'' points that are affected by the sensitive information.

\subsection{Analysis of Properties of Selected Data}
\begin{figure}[t]
    \centering
    \includegraphics[width=1\linewidth]{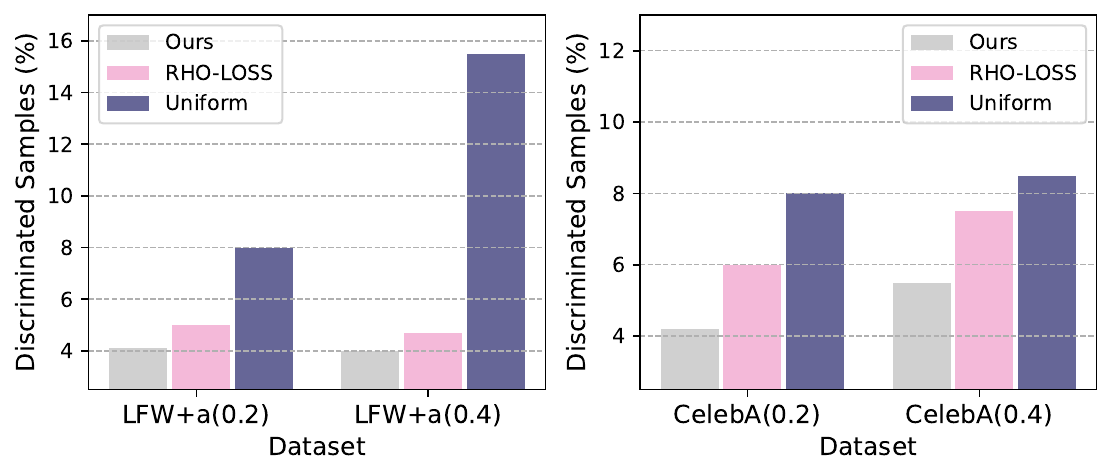}
    \caption{Proportion of selected instances discriminated by label bias using the proposed method (Ours), RHO-LOSS, and Uniform Sampling. The left plot corresponds to the LFW+a dataset, and the right plot corresponds to the CelebA dataset. Overall, we can observe that the proposed method has the lowest rate of discriminated sample selection.}
    \label{fig:selected_ratio}
\end{figure}

In this section, we analyze our proposed method by evaluating the proportion of fair instances to the selected data. \cref{fig:selected_ratio} reveal a pattern: as the amount of bias increases, the proportion of selected instances that are fair starts to decline due to the increased difficulty in distinguishing fair instances from unfair ones. However, our proposed selection method still maintains the highest ratio of selected fair instances, and significantly exceeds that of uniform sampling. This observation reinforces our method's superiority over other data selection methods in different label bias settings.

\begin{figure*}[t]
    \centering
    \includegraphics[width=1\linewidth]{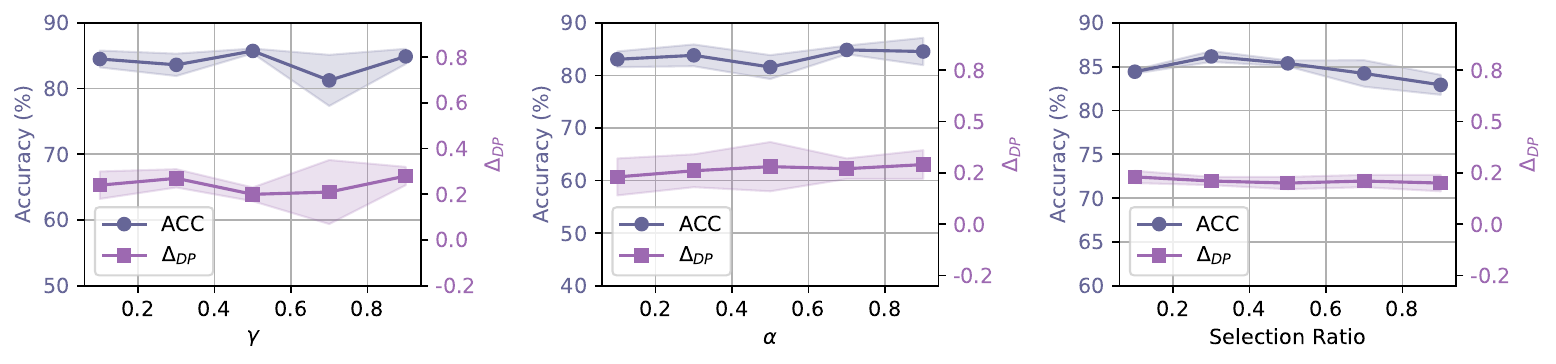}
    \caption{Ablation studies on critical hyperparameters, including $\gamma$, $\alpha$, and selection ratio, on the CelebA dataset with a 40\% label bias amount. We use blue to denote accuracy (left axis) and purple to denote fairness violation (right axis).}
    \label{fig:important_hyper}
\end{figure*}

\subsection{Ablation Studies}

In this section, we conduct ablation studies on zero-shot predictor, model architecture, and important hyperparameters. 

\textbf{Zero-shot Predictor.} In our experiments, we initially used CLIP-RN50 as the proxy model and have now extended testing to include ViT-B/16~\citep{dosovitskiy2021an} and the validation model from RHO-LOSS. Results in \cref{tab:variant_zero_shot_predictor} show consistent performance across different zero-shot backbones. For ViT-B/16 and RHO-LOSS models, baseline accuracies are 55\% and 65\% under 20\% label bias, and 52\% and 64\% under 40\% label bias, respectively. ViT-B/16 accuracies are slightly lower than RN50, but our data selection method with peer prediction maintains a comparable performance on CelebA at a 0.4 bias rate. RHO-LOSS model performance is slightly higher than RN50, but results are similar. This confirms the robustness of our method, which performs effectively regardless of the zero-shot predictor used.

\textbf{Backbone.} We test our proposed method with different Backbone architecture on CelebA dataset, including the variant of ResNet (ResNet-18 and ResNet-50) and the DenseNet-121~\cite{huang2017densely}. The results are displayed in \cref{tab:variant_zero_shot_predictor}.
From the results, we can see enhance model complexity generally improves both predictive performance and fairness, though these improvements are not significant. Overall, the impact of the model's structure on its performance is not obvious, which implies our proposed method is robust to the backbone structure.

\begin{table}[t]
    \centering
    \resizebox{\linewidth}{!}{
    \begin{tabular}{c c| c c| c c}
    \toprule
     \multicolumn{2}{c|}{Component}  & \multicolumn{2}{c|}{CelebA(0.2)} & \multicolumn{2}{c}{CelebA(0.4)}\\
     &   &  ACC & $\Delta_{\text{DP}}$ & ACC & $\Delta_{\text{DP}}$ \\
    \midrule
    \multirow{3}{*}{\makecell[c]{Zero-shot \\ Predictor}}  &  CLIP-RN50 & 85.5$_{\pm 0.6}$ & 0.21$_{\pm 0.02}$ & 85.2$_{\pm 1.7}$ & 0.23$_{\pm 0.06}$  \\
     &    ViT-B/16 &  86.4$_{\pm 0.2}$ & 0.20$_{\pm 0.03}$ & 84.5$_{\pm 1.2}$ & 0.19$_{\pm 0.02}$ \\
     &    R-V & 85.6$_{\pm 0.4}$ & 0.23$_{\pm 0.05}$ & 85.5$_{\pm 0.9}$ & 0.23$_{\pm 0.04}$\\
    \midrule
    \multirow{3}{*}{\makecell[c]{Backbone \\ Model}} & 
    ResNet-18 & 84.8$_{\pm 1.4}$ & 0.22$_{\pm 0.02}$ & 84.1$_{\pm 0.2}$ & 0.22$_{\pm 0.01}$\\
   & ResNet-50 & 86.5$_{\pm 0.6}$ & 0.21$_{\pm 0.02}$ & 85.2$_{\pm 1.7}$ & 0.20$_{\pm 0.01}$ \\
   & DenseNet-121 & 85.3$_{\pm 0.2}$ & 0.21$_{\pm 0.01}$ & 84.3$_{\pm 0.4}$ & 0.19$_{\pm 0.02}$
\\
     \bottomrule
    \end{tabular}}
    \caption{Test accuracy (\%) and fairness violations ($\Delta_{\text{DP}}$) for a variant zero-shot predictor (R-V denotes the validation model used in RHO-LOSS) and the backbone of our method.}
    \label{tab:variant_zero_shot_predictor}
\end{table}


\textbf{Hyperparameters.} We then analyze the effects of three hyperparameters: $\gamma$, $\alpha$, and the selection ratio.  In the first plot in \cref{fig:important_hyper}, we plot the difference of test accuracy and fairness violation with $\gamma$ varying, we set $\alpha = 0.1$. We can see that the accuracy slightly improves as $\gamma$ increases and the fairness violation appears downward trend as $\gamma$ increases. This align with the effect of $\gamma$ that controls the fairness level (fairness and accuracy should improve simultaneously when the data is unbiased~\citep{unlocking_fairness}).  In the second plot in \cref{fig:important_hyper}, we set $\gamma = 0.3$ and test different values of $\alpha$, we can see a similar trend for accuracy, but for the fairness violation, remains fairly stable. For the selection ratio, by default, is 10\% in the experiment. In the third plot in \cref{fig:important_hyper}, we plot the change of accuracy and fairness violation w.r.t. the selection ratio. We can see the accuracy increases when the selection ratio increases from 0.1 to 0.3 and then begins to drop. This demonstrates that increasing the selection ratio also increases the likelihood of picking data points that are not fair enough. However, due to the effect of the second term in \cref{eq: analysis_decomposition} acting as the fair regularizer, the fairness violation shows minimal variation.

\subsection{Convergence and Accuracy}
We conduct an experiment about convergence speed by testing the epochs required to reach target test accuracy on the two image datasets. Interestingly, due to our proposed data selection procedure, it has a faster convergence speed. In \cref{tab:epochs_converge}, we can see that the proposed method converges much faster than uniform sampling and RHO-LOSS. With fewer epochs, our proposed method can achieve the target accuracy point. In the meantime, as shown in \cref{tab:acc_compare}, the proposed method has the highest test accuracy compared to the other two methods. These results also align with \cref{fig:selected_ratio}, and can be explained as the proposed method is able to pick the fairest instances and therefore improve both accuracy and fairness at the same time. 

\begin{table}[t]
\centering
\resizebox{0.9\linewidth}{!}{
\begin{tabular}{@{}lccccc@{}}
\toprule
Dataset  & ACC & Uniform & RHO-LOSS & Ours   \\ \midrule
\multirow{2}{*}{LFW+a(0.2)} & 70\% & 41 & 32 & \textbf{21}\\
& 80\% & 73 & 65 & \textbf{55} \\
\multirow{2}{*}{LFW+a(0.4)} & 70\% & 77 & 69 & \textbf{23}  \\
& 80\% & - & 87 & \textbf{78}  \\ \midrule
\multirow{2}{*}{CelebA(0.2)} & 70\% & 65 & 33  & \textbf{25}   \\
& 80\% & 117 & 57  & \textbf{42}   \\
\multirow{2}{*}{CelebA(0.4)}  & 70\% & 94 & 46 &  \textbf{29}  \\
& 80\% & 125 & - & \textbf{85} \\
\bottomrule
\end{tabular}}
\caption{Epochs required to reach target test accuracy.}
\label{tab:epochs_converge}
\end{table}

\section{Related Work}
\label{sec:related_work}

\paragraph{Fair Learning with Label Bias} 
Fairness remains a critical and essential concern in real-world applications. A primary source of unfairness is label bias, which is typically modeled as fair (clean) labels being systematically flipped for individuals from certain demographic groups~\citep{unlocking_fairness}, i.e.,  if $S$ is a sensitive attribute and $Z$ a fair label, $P(Y=i \mid Z=j, S)$ occurs with $i \neq j$ and $i,j \in {0,1}$. A growing number of studies are exploring fair learning in settings with noisy labels to address this issue. For example, \citet{gpl} applied group-dependent label noise and derived fairness constraints on corrupted data. \citet{pmlr-v108-jiang20a} propose a re-weighting method to correct instances affected by label bias. Building on these, \citet{Dai2020LabelBL} presents a framework to understand the combined effects of label bias and data distribution shifts in the context of fairness from a fundamental perspective. These works share a common framework that considers the noise rate of labels to improve the robustness of fair learning methods. However, these approaches involve modifying models and intervening in the training process, which limits their flexibility for complex systems and large-scale datasets. To overcome this limitation, we adopt a data selection framework that enhances both flexibility and efficiency, making it suitable for practical applications.

\paragraph{Data Selection Methods}
As data sizes increase, using all available data for training becomes inefficient. Therefore, data selection methods have been developed to selectively train on only the most useful data, thereby enhancing efficiency and reducing computational costs. Previous approaches like curriculum learning~\citep{Curriculum/10.1145/1553374.1553380} typically choose data points from easy to hard. This method can lead to redundancy, as once such data points are learned, they should not be learned again. Other methods select data based on high training loss or high prediction uncertainty~\citep{Loshchilov2017DecoupledWD,Kawaguchi2019OrderedSA,DBLP:journals/corr/jiang,DBLP:journals/corr/coleman,DBLP:journals/corr/LoshchilovH15}. A common issue with these approaches is their tendency to choose outliers or noisy points when focusing on high loss or uncertainty. To address this issue, new selection methods that assess the data's impact on generalization loss, derived from a holdout set~\citep{Killamsetty2020GLISTERGB,pmlr-v162-mindermann22a}, have been developed. Addressing the problem of less principled approximations in these methods and circumventing the need for a holdout set, \citet{deng2023towards} implemented a Bayesian approach, enhancing the validity of approximations and eliminating the reliance on a holdout set through the use of a zero-shot predictor. Inspired by this work, we also utilize the peer prediction mechanism to ensure the fairness of the zero-shot predictor.


\section{Conclusions}
This paper addresses label bias in fairness for large-scale datasets by proposing a fair data selection strategy that aligns the RHO-LOSS criterion with a fair distribution and uses a zero-shot predictor to eliminate the need for a clean holdout set. The approach enhances model fairness and accuracy while ensuring the selected data better represents a fair distribution through peer prediction mechanism, suitable for training robust models across various real-world applications. The proposed selection method is constructed from three aspects: (1) deriving a tractable selection function to pick data less affected by label bias, (2) eliminating the need for an additional holdout set previously required for validation, and (3) incorporating a peer prediction mechanism to ensure the fairness of the validation model. Experiments demonstrate the method’s effectiveness in mitigating label bias, achieving faster convergence, and higher test accuracy compared to alternatives. While this work primarily focuses on fairness within the defined scope, future research will explore more complex settings, including out-of-distribution (OOD) scenarios and the role of confounders in fairness.

\section{Acknowledgments}
We sincerely thank the reviewers for their careful reading of the manuscript and their valuable suggestions. This work was supported by the NSFC Project (No. 62106121), the MOE Project of Key Research Institute of Humanities and Social Sciences (22JJD110001), the fundamental research funds for the central universities, and the research funds of Renmin University of China (24XNKJ13). 

\bibliography{aaai25}

\clearpage
\appendix
\onecolumn

\section*{Reproducibility Checklist}

\subsection*{1. This paper:}
\begin{itemize}
    \item Includes a conceptual outline and/or pseudocode description of AI methods introduced (yes)
    \item Clearly delineates statements that are opinions, hypothesis, and speculation from objective facts and results (yes)
    \item Provides well marked pedagogical references for less-familiare readers to gain background necessary to replicate the paper (yes)
\end{itemize}

\subsection*{2. Does this paper make theoretical contributions? (yes)}

\subsubsection*{2.1 If yes, please complete the list below.}
\begin{itemize}
    \item All assumptions and restrictions are stated clearly and formally. (yes)
    \item All novel claims are stated formally (e.g., in theorem statements). (yes)
    \item Proofs of all novel claims are included. (yes)
    \item Proof sketches or intuitions are given for complex and/or novel results. (yes)
    \item Appropriate citations to theoretical tools used are given. (yes)
    \item All theoretical claims are demonstrated empirically to hold. (yes)
    \item All experimental code used to eliminate or disprove claims is included. (NA)  
\end{itemize}

\subsection*{3. Does this paper rely on one or more datasets? (yes)}

\subsubsection{3.1 If yes, please complete the list below.}
\begin{itemize}
    \item  A motivation is given for why the experiments are conducted on the selected datasets (NA)
    \item All novel datasets introduced in this paper are included in a data appendix. (NA)
    \item All novel datasets introduced in this paper will be made publicly available upon publication of the paper with a license that allows free usage for research purposes. (NA)
    \item All datasets drawn from the existing literature (potentially including authors’ own previously published work) are accompanied by appropriate citations. (yes)
    \item All datasets drawn from the existing literature (potentially including authors’ own previously published work) are publicly available. (yes)
    \item All datasets that are not publicly available are described in detail, with explanation why publicly available alternatives are not scientifically satisficing. (NA)
\end{itemize}

\subsection*{4. Does this paper include computational experiments? (yes)}

\subsubsection*{If yes, please complete the list below.}
\begin{itemize}
    \item  Any code required for pre-processing data is included in the appendix. (yes).
    \item All source code required for conducting and analyzing the experiments is included in a code appendix. (yes)
    \item All source code required for conducting and analyzing the experiments will be made publicly available upon publication of the paper with a license that allows free usage for research purposes. (yes)
    \item All source code implementing new methods have comments detailing the implementation, with references to the paper where each step comes from (yes)
    \item If an algorithm depends on randomness, then the method used for setting seeds is described in a way sufficient to allow replication of results. (yes)
    \item This paper specifies the computing infrastructure used for running experiments (hardware and software), including GPU/CPU models; amount of memory; operating system; names and versions of relevant software libraries and frameworks. (yes)
    \item This paper formally describes evaluation metrics used and explains the motivation for choosing these metrics. (yes)
    \item This paper states the number of algorithm runs used to compute each reported result. (yes)
    \item Analysis of experiments goes beyond single-dimensional summaries of performance (e.g., average; median) to include measures of variation, confidence, or other distributional information. (yes)
    \item The significance of any improvement or decrease in performance is judged using appropriate statistical tests (e.g., Wilcoxon signed-rank). (no)
    \item This paper lists all final (hyper-)parameters used for each model/algorithm in the paper’s experiments. (yes)
    \item This paper states the number and range of values tried per (hyper-) parameter during development of the paper, along with the criterion used for selecting the final parameter setting. (partial)
\end{itemize}

\section{Refining Selection Function}
In this section, we detail the derivation of equations related to the new optimization goal, which focuses on refining the selection function to consider fair distribution.

\subsection{Derivation of the optimization problem in Equation 5}

\begin{equation*}
\begin{aligned}
    &\underset{(x,y)\in B_{t+1}}{\text{max}}\frac{1}{m}\sum_{i=1}^m \sum_{s\in [S]} \bar{p}(s)[\log p(y^*_i\mid x^*_i,\mathcal{D}_t \cup (x,y),s)]\\
    & \Leftrightarrow  \underset{(x,y)\in B_{t+1}}{\text{max}}\sum_{s\in [S]} \bar{p}(s)[\log p(\mathcal{D}^* \mid \mathcal{D}_t \cup (x,y),s)]. 
\end{aligned}
\end{equation*}

\subsection{Derivation of the Bayes' rule in Equation 6}
\begin{equation*}
    \begin{aligned}
        p(\mathcal{D}^*\mid \mathcal{D}_t\cup (x,y),s) &= \frac{p(y\mid x,\mathcal{D}^*, \mathcal{D}_{t},s)}{p(y\mid x, \mathcal{D}_{t},s)}\cdot p(\mathcal{D}^* \mid \mathcal{D}_{t},x,s)\\
        &= \frac{p(y\mid x,\mathcal{D}^*, \mathcal{D}_{t},s)}{p(y\mid x, \mathcal{D}_{t},s)}\cdot p(\mathcal{D}^* \mid \mathcal{D}_{t}).
    \end{aligned}
\end{equation*}

\subsection{Derivation of the lower bound in Equation 8}
\label{sec: appendix_lowerbound}
\begin{equation*}
\begin{aligned}
    &\log p(y\mid x,\mathcal{D}^*,\mathcal{D}_t,s) \\
    =& \log \int p(\theta\mid \mathcal{D}^*,\mathcal{D}_t)p(y\mid x,\theta,s)d\theta \\ 
    =& \log \int \frac{p(\mathcal{D}_t\mid \theta)p(\theta\mid \mathcal{D}^*)}{p(\mathcal{D}_t\mid \mathcal{D}^*)}p(y\mid x,\theta,s)d\theta \\ 
    =& \log \int p(\mathcal{D}_t \mid \theta)p(\theta \mid \mathcal{D}^*)p(y\mid x,\theta,s)d\theta - \log p(\mathcal{D}_t\mid \mathcal{D}^*) \\ 
    \ge & \int p(\theta \mid \mathcal{D}^*) \log [p(\mathcal{D}_t\mid\theta)p(y\mid x,\theta,s)]d\theta - \log p(\mathcal{D}_t\mid \mathcal{D}^*)\\ 
    =& \mathbb{E}_{p(\theta\mid \mathcal{D}^*)}\log p(\mathcal{D}_t\mid \theta)+\mathbb{E}_{p(\theta\mid \mathcal{D}^*)}\log p(y\mid x,\theta,s)-\log p(\mathcal{D}_t\mid \mathcal{D}^*) 
\end{aligned}
\end{equation*}
Drop the terms irrelevant to $(x,y)$, we get:
\begin{equation*}
    \log p(y\mid x,\mathcal{D}^*,\mathcal{D}_t,s) \ge \mathbb{E}_{p(\theta \mid \mathcal{D}^*)}\log p(y\mid x,\theta,s).
\end{equation*}

\section{Derivation of the equivalent expectation version of the peer loss in Equation 12}
\label{sec: appendix_derivation_expectation}
We denote $\mathcal{D}_s$ as a shorthand for $\mathcal{D}_{S=s}$ for $s\in [S]$. For each $x_i$ belonging to the demographic group $S=s$, let $|\mathcal{D}_s| = N_s$, and $[I]$ be the set of sample index in $\mathcal{D}_s$. By taking the expectation over $(X_{i_1},Y_{i_2})$, we have:

\begin{equation*}
\begin{aligned}
    & \frac{1}{N_s}\sum_{i\in [I]]} \mathbb{E}_{X_{i_1}\mid \mathcal{D}_{s}, Y_{i_2}\mid \mathcal{D}_{s'}}\left[L[y_i,\Tilde{f}(x_i),s]-\gamma L[Y_{i_2},\Tilde{f}(X_{i_1}),s]\right ] \\
    &= \frac{1}{N_s}\sum_{i \in [I]} \left [L[y_i,\Tilde{f}(x_i),s] -\sum_{i' \in [I], i'\neq i} P(X_{i_1}=x_{i'}\mid S=s)\mathbb{E}_{Y\mid \mathcal{D}_{s'}}\gamma L[Y,\Tilde{f}(x_{i'}),s]\right ] \\
    &= \frac{1}{N_s}\sum_{i \in [I]} \left [L[y_i,\Tilde{f}(x_i),s] -\sum_{i' \in [I], i' \neq i} \frac{1}{N_s}\mathbb{E}_{Y\mid \mathcal{D}_{s'}}\gamma L[Y,\Tilde{f}(x_{i'}),s]\right ]\\
    &=\frac{1}{N_s}\sum_{i \in [I]} \left [L[y_i,\Tilde{f}(x_i),s] -\mathbb{E}_{Y\mid \mathcal{D}_{s'}}\gamma L[Y,\Tilde{f}(x_{i}),s]\right ]
\end{aligned}
\end{equation*}
So, the result of the fairness-regularized loss on the zero-shot predictor will be:
\begin{equation*}
    L[y,\Tilde{f}(x),s] -\mathbb{E}_{Y\mid \mathcal{D}_{s'}}\gamma L[Y,\Tilde{f}(x),s]
\end{equation*}
Is is worth noting that the last term is taking the expectation of $Y$ over $s'$, while $x$ is drawn from $s$, we designed this intentionally: in the peer prediction mechanism, we sample $x_{i_1}$ from $D_s$ and $y_{i_2}$ from $D_{s'}$. 

\section{Decompose the Loss}
\label{sec:appendix_decomposition}
In this section, we demonstrate the effectiveness of the loss proposed in Equation 12 by decomposing its expected version.
\begin{equation}
\begin{aligned}
\label{eq:appendix_peer_loss_goal}
   \underbrace{L[y,\Tilde{f}(x),s]}_\text{A}- \underbrace{\gamma \mathbb{E}_{Y\mid \mathcal{D}_{s'}}[L[Y, \Tilde{f}(x),s]]}_\text{B}.
    \end{aligned}
\end{equation}

\subsection{Decomposition of A}
To clarify the notation, we use $\Tilde{D}$ to denote the observed distribution, which contains label bias, and $D$ to denote the underlying clean fair distribution. In the context of label bias, the flip does not depend on $X$; that is, $P(Y=j\mid Z=i, S=s, X) = P(Y=j\mid Z=i, S=s)$. For brevity, we refer to this simply as $T_{ij}^s$. We use $\theta_s^+ = P(Y=1\mid Z=0, S=s)$ and $\theta_s^- = P(Y=0\mid Z=1,S=s)$ to denote the specific flip rate. The following formulation incorporates the independence between $P(S)$ and $P(Z)$. For $Y$, $Z$ and $S$, we only consider the binary case and we use $K$ to denote the number of class ($[K]=\{0,1\}$ in our case). Part A in \cref{eq:appendix_peer_loss_goal} can be decomposed into: 

\begin{equation}
    \begin{aligned}
        & \mathbb{E}_{\Tilde{D}}\left [L[Y,\Tilde{f}(X),S]\right ] \\
        &= \sum_{j\in [K]} \sum_{i \in [K]} \sum_{s \in [S]} P(Z=i)P(S=s)\mathbb{E}_{D\mid Z=i, S=s}\left [T^s_{ij}L[j,\Tilde{f}(X),s] \right ]\\
        &= \sum_{j\in [K]} \sum_{i \in [K]} \sum_{s \in [S]}P(Z=i)P(S=s)\mathbb{E}_{D\mid Z=i,S=s}T_{ij}^s\mathbb{E}_{D\mid Z=i, S=s}\left[L[j,\Tilde{f}(X),s]\right]\\
        &+ \sum_{j\in [K]} \sum_{i \in [K]} \sum_{s \in [S]}P(Z=i)P(S=s)\text{Cov}_{D\mid Z=i,S=s}\left ( T_{ij}^s , L[j,\Tilde{f}(X),s]\right )
    \end{aligned}
\end{equation}

The first term can be further decomposed into:
{\small
\begin{equation}
    \begin{aligned}
       & \underbrace{\sum_{s \in [S]}P(S=s)\Big[\mathbb{E}_D(1-\theta_s^- - \theta_s^+)\mathbb{E}_{D\mid S=s}L[Z,\Tilde{f}(X),s]}_\text{Term 1} \\
       &+ \underbrace{P(Z=1)\mathbb{E}_{D\mid Z=1,S=s}\theta_s^+\mathbb{E}_{D\mid Z=1,S=s}L[1,\Tilde{f}(X),s]+P(Z=0)\mathbb{E}_{D\mid Z=0,S=s}\theta_s^-\mathbb{E}_{D\mid Z=0,S=s}L[0,\Tilde{f}(X),s]}_\text{Term 2}\\
       &+ \underbrace{\sum_{j\in [K],i\in [K],i\neq j}P(Z=i)\mathbb{E}_{D\mid Z=i, S=s}T_{ij}^s\mathbb{E}_{D\mid Z=i, S=s}L[j,\Tilde{f}(X),s]\Big]}_\text{Term 3}\\
    \end{aligned}
\end{equation}}
The second term will be:
{\small
\begin{equation}
\label{eq:appendix_second_term}
    \begin{aligned}
         &\sum_{s\in [S]} P(S=s) \left[ \sum_{j\in [K]} P(Z=j) \mathbb{E}_{D\mid Z=j,S=s}\big (T_{jj}^s-\bar{T}{_{jj}^s}\big ) \right . 
 \left(L[j,\Tilde{f}(X),s]- \mathbb{E}_{D\mid Z=j,S=s}[L[j, \Tilde{f}(X),s]\right) \\
    &+  \left.\sum_{j\in [K]} \sum_{i \in [K], i\neq j} P(Z=i) \mathbb{E}_{D\mid Z=i,S=s}\left(T_{ij}^s-\bar{T}{_{ij}^s}\right) \times \left(L[j,\Tilde{f}(X),s]- \mathbb{E}_{D \mid Z=i,S=s}L[j,\Tilde{f}(X),s]\right)\right],
    \end{aligned}
\end{equation}}
where $\bar{T}{_{ij}^s} = \mathbb{E}_{D\mid Z=i,S=s}T_{ij}^s$. Then, combine \cref{eq:appendix_second_term} with Term 3, we get:
{\small
\begin{equation}
\label{eq:appendix_combine_term3}
    \begin{aligned}
    & \sum_{s\in [S]} P(S=s)\left [P(Z=1)\mathbb{E}_{D \mid Z=1,S=s}\left(1-\theta_s^{-}-\mathbb{E}_{D\mid Z=1,S=s}(1-\theta_s^-)\right)L(1,\Tilde{f}(x),s]\right.\\
    &+P(Z=0)\mathbb{E}_{D \mid Z=0,S=s}\left(1-\theta_s^{+}-\mathbb{E}_{D \mid Z=0,S=s}(1-\theta_s^+)\right)L[0,\Tilde{f}(X),s]\\
    &\left . +P(Z=0)\mathbb{E}_{D\mid Z=0,S=s}(\theta_s^{+}\ell[1,\Tilde{f}(X)])+P(Z=1)\mathbb{E}_{D \mid Z=1,S=s}(\theta_s^{-}L[0,\Tilde{f}(X),s])\right].
    \end{aligned}
\end{equation}}

Integrating \cref{eq:appendix_combine_term3} with Term 1 and Term 2, and considering that the fair and clean distribution $D$ remains unaffected by the sensitive variable $S$, we obtain:
\begin{equation}
\label{eq:appendix_decompose_first_part}
    \begin{aligned}
    &\mathbb{E}_{\Tilde{D}}[L[Y,\Tilde{f}(X),S]]\\
    &=  \mathbb{E}_{D}\Delta_s L[Z,\Tilde{f}(X),S]+\sum_{s\in [S]} \sum_{j \in [K]}\sum_{i\in [K]}P(S=s)P(Z=i)\mathbb{E}_{D\mid Z=i,S=s}U_{ij}^s L[j,\Tilde{f}(X),s]
    \end{aligned}
\end{equation}
where
 \begin{equation*}
  U_{ij}^s = \begin{cases}
        T_{ij}^s \text{     if $i \neq j$,}\\
        T_{jj}^s - \bar{T}{_{jj}^s} \text{    if $i=j$}.
        \end{cases}
 \end{equation*}

\subsection{Decomposition of B}
\begin{equation}
\label{eq: appendix_decomposition_b}
    \begin{aligned}
        &\mathbb{E}_{\Tilde{D}}\left[-\gamma \mathbb{E}_{Y\mid \Tilde{D},S=s}L\left[Y,\Tilde{f}(X),S=s'\right] \right]\\
        &= -\gamma \int_{\Tilde{D}}\left[P(\Tilde{D}\mid S=0)P(S=0)\sum_{j \in [K]} P(Y=j\mid \Tilde{D},S=0)L\left[Y=j,\Tilde{f}(X),S=1\right]\right .\\
        &\qquad +  \left . P(\Tilde{D}\mid S=1)P(S=1)\sum_{j\in [K]} P(Y=j\mid \Tilde{D},S=0)L\left[Y=j,\Tilde{f}(X),S=1\right] \right ]\\
        &= -\gamma \sum_{j \in [K]} \left[P(Y=j,S=0)\mathbb{E}_{D_{X\mid S=1}}L[j,\Tilde{f}(X),S=1]+P(Y=j,S=1)\mathbb{E}_{D_{X\mid S=0}}L[j,\Tilde{f}(X),S=0]\right]\\
        &=-\gamma \sum_{j \in [K]} \left[P(Y=j,S=0)\left(\mathbb{E}_{D_{X\mid S=1}}L[j,\Tilde{f}(X),S=1]-\mathbb{E}_{D_{X\mid S=0}}L[j,\Tilde{f}(X),S=0]\right)\right.\\
        & \qquad + P(Y=j,S=1)\left(\mathbb{E}_{D_{X\mid S=0}}L[j,\Tilde{f}(X),S=0]-\mathbb{E}_{D_{X\mid S=1}}L[j,\Tilde{f}(X),S=1]\right)\\
        &\qquad \left.+P(Y=j,S=1)\mathbb{E}_{D_{X\mid S=1}}L[j,\Tilde{f}(X),S=1]+ P(Y=j,S=0)\mathbb{E}_{D_{X\mid S=0}}L[j,\Tilde{f}(X),S=0]\right ]\\
        &= -\gamma \sum_{j\in [K]} d_j^s\left(\mathbb{E}_{D_{X\mid S=1}}L[j,\Tilde{f}(X),S=1]-\mathbb{E}_{D_{X\mid S=0}}L[j,\Tilde{f}(X),S=0]\right) \\
        & \qquad -\gamma \sum_{j\in [K]}\sum_S P(Y=j,S=s)\mathbb{E}_{D_{X\mid S=s}}L[j,\Tilde{f}(X),s],\\
    \end{aligned}
\end{equation}
where $d^s_j = \left(P(Y=j,S=0)-P(Y=j,S=1)\right)$, which measures the discrepancy in the probability of observed label $j$ between two demographic groups. For the second term in \cref{eq: appendix_decomposition_b}, we can further expand it as:
\begin{equation}
\label{eq:appendix_second_term_fair}
\begin{aligned}
    &-\gamma \sum_{j\in [K]} \sum_{s\in [S]} P(Y=j,S=s)\mathbb{E}_{D_{X\mid S=s}}L[j,\Tilde{f}(X),s]\\
    &= -\sum_{j\in [K]} \sum_{i \in [K]} \sum_{s\in [S]} P(Z=i)\mathbb{E}_{D\mid Z=i, S=s}\left[\gamma P(Y=j,S=s) L[j,\Tilde{f}(X),s]\right]
\end{aligned}
\end{equation}

Combine \cref{eq:appendix_second_term_fair} with the second term in \cref{eq:appendix_decompose_first_part}, we get:

\begin{equation}
\label{eq:appendix_second_part_b_}
    \sum_{j\in [K]} \sum_{i \in [K]} \sum_{s\in [S]} P(S=s)P(Z=i)\mathbb{E}_{D\mid Z=i, S=s}[(U_{ij}^s - \gamma P(Y=j\mid S=s) )L[j,\Tilde{f}(X),s]]
\end{equation}

Finally, the original expected loss is decomposed into the following components:

\begin{equation}
\begin{aligned}
  &\mathbb{E}_{\Tilde{D}}\left [L[Z,\Tilde{f}(X),S]- \gamma \mathbb{E}_{Z\mid \mathcal{D}_{s'}}L[Z, \Tilde{f}(x),s]\right ]\\
   &= \underbrace{\mathbb{E}_{D}\Delta_s L[Z,\Tilde{f}(X),S]}_\text{fair model}\\
   &+ \underbrace{\sum_{j\in [K]} \sum_{i \in [K]} \sum_{s\in [S]} P(S=s)P(Z=i)\mathbb{E}_{D\mid Z=i, S=s}[(U_{ij}^s - \gamma P(Y=j\mid S=s) )L[j,\Tilde{f}(X),s]]}_\text{Penalty on noisy loss}\\
   &\underbrace{-\gamma \sum_{j\in [K]}  p_j^s \left [\mathbb{E}_{D_{X\mid S=1}}L[j,\Tilde{f}(X)]-\mathbb{E}_{D_{X\mid S=0}}L[j,\Tilde{f}(X)]\right ],}_\text{Penalty for the disagreement between demographic groups}
    \end{aligned}
\end{equation}
where
 \begin{equation*}
 \begin{aligned}
     \Delta_s &= 1-\theta_s^- - \theta_s^+\\
 p_j^s &= P(Y=j, S=0) - P(Y=j, S=1)\\
  U_{ij}^s &= \begin{cases}
        T_{ij}^s \text{     if $i \neq j$,}\\
        T_{jj}^s - \mathbb{E}_{D\mid Y=j,S=s}T_{jj}^s \text{    if $i=j$}.
        \end{cases}
    \end{aligned}
 \end{equation*}

\end{document}